# Asymmetric Separation for Local Independence Graphs

**Vanessa Didelez**
Department of Statistical Science
University College London, UK
vanessa@stats.ucl.ac.uk

## Abstract

Directed possibly cyclic graphs have been proposed by Didelez (2000) and Nodelmann et al. (2002) in order to represent the dynamic dependencies among stochastic processes. These dependencies are based on a generalization of Granger–causality to continuous time, first developed by Schweder (1970) for Markov processes, who called them *local dependencies*. They deserve special attention as they are asymmetric. In this paper we focus on their graphical representation and develop an asymmetric notion of separation. The properties of this graph separation as well as local independence are investigated in detail within a framework of asymmetric (semi)graphoids allowing insight into what information can be read off these graphs.

## 1 INTRODUCTION

Classical graphical models and Bayesian networks represent (conditional) independence between random variables. They can be adapted to variables observed in discrete or discretized time, like e.g. time series or dynamic Bayesian networks. This does not, however, capture the intuitive notion of dynamic dependence between processes, which is asymmetric, and we therefore propose an alternative.

Put informally, we are interested in the following kind of dynamic (conditional) independencies among stochastic processes $X(t), Y(t)$ and $Z(t)$:

present of $X \perp\!\!\!\perp$ past of $Y$ | past of $(X, Z)$,

which will be denoted by $Y \not\to X | Z$, or more formally

$$X(t) \perp\!\!\!\perp \mathcal{F}^Y_{t^-} \mid \mathcal{F}^{X,Z}_{t^-},$$

where $\mathcal{F}_{t^-}$ is the history of a process. Such independence underlies the notion of Granger non–causality for time series (Granger, 1969), which has been used as the basis for a graphical representation by Eichler (2002). It also underlies the continuous–time notion of *local independence* given by Schweder (1970) for Markov processes, and its bivariate version for general continuous–time processes by Aalen (1987) with a multivariate version in Didelez (2000, pp.65). A special case of Schweder's concept are the dependencies represented in the continuous time Bayesian networks developed by Nodelman et al. (2002) for homogenous Markov processes. Cyclic graphs have also been proposed to represent non–recursive structural equation models or feedback processes (Sprites, 1995; Pearl and Dechter, 1996), but with the above notion of dependence between past and present we explicitly consider processes instead of cross sectional measurements of variables in equilibrium.

Consider the following example illustrating the idea of local dependence and its graphical representation. In some countries programs exist to assist the elderly, e.g. through regular visits by nurses. Assume that these visits reduce the frequency of hospitalizations but do not in any other way affect future survival (this could be the null hypothesis). Assume further that the frequency of visits is intensified when a person has previously been hospitalized. Also assume that the general health status affects the rate of future hospitalization and the survival but not the frequency of future visits. The proposed graphical representation of these assumptions is given in Figure 1.

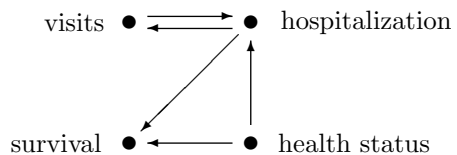

Figure 1: Example for local independence graph.

The outline of the paper is as follows. We first review, in Section 2, the notion of local independence for Markov processes and its generalization to multi-state processes. A graphical representation as well as a suitable asymmetric separation are proposed in Section 3. The properties of local independence and the corresponding graphical representation are investigated in Section 4 within the framework of asymmetric graphoids.

## 2 LOCAL INDEPENDENCE

### 2.1 MARKOV PROCESSES

We consider a first order Markov process $\mathbf{Y}(t)$, $t \in \mathcal{T}$, with a finite state space $\mathcal{S}$ and with transition intensities $\alpha_{qr}(t)$, $q \neq r \in \mathcal{S}$, which are assumed to exist. The Markov process $\mathbf{Y}$ is assumed to consist of *components* in the following sense (Schweder, 1970).

**Definition 2.1** *Composable Markov process*
Let $V = \{1, \ldots, K\}$, $K \geq 2$, and assume that there are $K$ spaces $\mathcal{S}_k, k \in V$, with $|\mathcal{S}_k| \geq 2$, and that there exists a one-to-one mapping $f$ of $\mathcal{S}$ onto $\bigotimes_{k \in V} \mathcal{S}_k$ so that elements $\mathbf{y} \in \mathcal{S}$ can be identified with elements $(y_1, \ldots, y_K) \in \bigotimes_{k \in V} \mathcal{S}_k$. Then, a Markov process $\mathbf{Y}$ is a composable process with components $Y_1, \ldots, Y_K$ given by $f(\mathbf{Y}(t)) = (Y_1(t), \ldots, Y_K(t))$ if for all $A \subset V$, $|A| \geq 2$,

$$\lim_{h \downarrow 0} \frac{1}{h} P\left( \bigcap_{k \in A} \{Y_k(t+h) \neq y_k\} \middle| \bigcap_{k \in A} \{Y_k(t) = y_k\} \right) = 0$$

for all $y_k \in \mathcal{S}_k, k \in V$, and $t \in \mathcal{T}$. We then write $\mathbf{Y} \sim (Y_1, \ldots, Y_K)$.

The definition implies that for a composable process the probability that more than one component changes in a short period of length $h$ is of magnitude $o(h)$. Hence, any change of state can be represented as a change in only one of the components which justifies regarding the processes as *composed* of different components. The components are not necessarily unique. If for example $\mathbf{Y} \sim (Y_1, \ldots, Y_K)$ then $\mathbf{Y} \sim (\mathbf{Y}_A, \mathbf{Y}_B)$ with $A \subset V$ and $B = V \setminus A$. Note that the components of such a Markov process correspond to the *local variables* of Nodelman et al. (2002).

From now, we consider composable finite Markov processes (CFMPs) and write $\alpha(t; (\mathbf{y}, \mathbf{y}'))$ instead of $\alpha_{\mathbf{y}\mathbf{y}'}(t)$ for notational convenience. Schweder (1970) proves the following.

**Corollary 2.2** *Transition intensities for CFMPs*
Let $\mathbf{Y} \sim (Y_1, \ldots, Y_K)$ be a CFMP. The intensity $\alpha(t; (\mathbf{y}, \mathbf{y}'))$ for any $\mathbf{y} \neq \mathbf{y}' \in \mathcal{S}$ is given by

$$\alpha(t; (\mathbf{y}, \mathbf{y}')) = \begin{cases} \alpha_k(t; (\mathbf{y}, y'_k)), & y_k \neq y'_k \wedge \mathbf{y}_{-k} = \mathbf{y}'_{-k} \\ 0, & \text{else}, \end{cases}$$

where $\mathbf{y}_{-k} = \mathbf{y}_{V \setminus \{k\}}$, and

$$\alpha_k(t; (\mathbf{y}, y'_k)) = \lim_{h \downarrow 0} \frac{1}{h} P\left(Y_k(t+h) = y'_k \mid \mathbf{Y}(t) = \mathbf{y}\right).$$

The dependence structure of $(Y_1, \ldots, Y_K)$ is thus determined by the quantities $\alpha_k(t; (\mathbf{y}, y'_k)), \mathbf{y} \in \mathcal{S}, y'_k \in \mathcal{S}_k \setminus \{y_k\}, k \in V$.

**Definition 2.3** *Local independence in a CFMP*
Let $\mathbf{Y} \sim (Y_1, \ldots, Y_K)$ be a CFMP. Then, $Y_j$ is locally independent of $Y_k$, $k \neq j$, if and only if $\alpha_j(t; (\mathbf{y}, y'_j))$ is constant in the $k$-th component $y_k$ of the first argument $\forall\, \mathbf{y}_{-k} \in \mathcal{S}_{-k}$ and $y'_j \in \mathcal{S}_j$, $y'_j \neq y_j$, for all $t \in \mathcal{T}$. We denote this by $\{k\} \not\rightarrow \{j\} | V \setminus \{j, k\}$.
For disjoint subsets $A, B, C$ of $V$ we say that $B \not\rightarrow A | C$ if for all $j \in A$ the transition intensities $\alpha_j(t; (\mathbf{y}, y'_j))$ are constant in all the components $\mathbf{y}_B$ of $\mathbf{y}$ for any $\mathbf{y}_{A \cup C} \in \mathcal{S}_{A \cup C}$ and $y'_j \in \mathcal{S}_j$, $y'_j \neq y_j$, written as $B \not\rightarrow A | C$.

Since for small $h > 0$

$$P\left(Y_j(t+h) = y'_j \mid \mathbf{Y}(t) = \mathbf{y}\right) \approx h \cdot \alpha_j(t; (\mathbf{y}, y'_j))$$

one can roughly say that $\{k\} \not\rightarrow \{j\} | V \setminus \{j, k\}$ means $Y_j(t + h)$ is conditionally independent of $Y_k(t)$ given $\mathbf{Y}_{-k}(t)$ for small $h$, hence *local* independence.

A straightforward graphical representation of local dependence structures is to depict components $Y_k$ by nodes and local dependencies by directed edges. We can then summarize the transition intensities as $\alpha_k(t; (\mathbf{y}_{\text{cl}(k)}, y'_k))$, where $\text{cl}(k)$ is the closure of node $k$, i.e. its parents and itself. In the terminology of Nodelman et al. (2002), a component $Y_k$ can be regarded as a *conditional* Markov process, the transition intensities of which depend on the states of $\mathbf{Y}_{\text{pa}(k)}$ and can be represented in a collection of conditional transition matrices for the transitions $y_k \rightarrow y'_k$, $y_k \neq y'_k \in \mathcal{S}_k$, one matrix for each value of $\mathbf{y}_{\text{pa}(k)} \in \mathcal{S}_{\text{pa}(k)}$.

In the above we assume that a component is always locally dependent on its own past so that this is not explicitly stated in the conditioning set when writing $\{k\} \not\rightarrow \{j\} | V \setminus \{j, k\}$ or in the parent set in the corresponding graph. For practical applications it seems unlikely that a component is locally independent of itself, but such processes can of course be constructed. If we wanted to allow this, the graphs would need to be extended to include self–loops $(j, j)$, $j \in V$, and there would be a difference between $\{k\} \not\rightarrow \{j\} | V \setminus \{j, k\}$ and $\{k\} \not\rightarrow \{j\} | V \setminus \{k\}$. But we do not pursue this any further here.

## 2.2 MULTI–STATE PROCESSES

Marginally, a subprocess $\mathbf{Y}_A$, $A \subset V$, of a CFMP is not necessarily Markov anymore unless $\mathrm{pa}(A) = \emptyset$ (Didelez, 2005b). Hence it is useful to generalize the notion of local independence to a larger class of processes like multi–state processes that are not necessarily Markov (cf. Didelez, 2005a).

Let $\mathbf{Y}$ denote a multi–state process with a one–to–one mapping of its state space $\mathcal{S}$ onto $\bigotimes \mathcal{S}_k$ yielding the components $Y_1, \ldots, Y_K$ similar to Definition 2.1. Let further $\mathcal{F}_t^A$, $A \subset V$, denote the filtration generated by the subprocess $\mathbf{Y}_A$, i.e. $\mathcal{F}_t^A = \sigma\{\mathbf{Y}_A(s), s \leq t\}$. Then we assume that for each component $Y_k$ there exists a process $\Lambda_k$ such that $M_k = Y_k - \Lambda_k$ is a $\mathcal{F}_t^V$–martingale, i.e.

$$Y_k = \Lambda_k + M_k. \tag{1}$$

Property (1) is the Doob–Meyer decomposition (cf. Andersen et al., 1993) and is similar to a regression model, where $\Lambda_k(t)$ is the predictor based on the past $\mathcal{F}_{t-}^V$ of this *and all other components* (formally this means that $\Lambda_k(t)$ is a $\mathcal{F}_t^V$–predictable compensator), and $M_t$ corresponds to a zero–mean error term (conditionally on the past). Further, the martingales $M_k$, $k = 1, \ldots, K$, are assumed to be orthogonal, which can be regarded as a dynamic version of an 'independent errors' assumption. It also ensures that no two components 'jump' at the same time in analogy to Definition 2.1.

For a Markov process the predictor $\Lambda_k(t)$ can be constructed from the transition intensities and suitable indicator functions for the state of a process just before the transition. If it is absolutely continuous then its derivative $\lambda_k(t)$ is called the intensity process. E.g. for a Markov process $Y$, and considering one particular transition $q \to r$, we have $\lambda_{qr}(t) = \alpha_{qr}(t) I_{\{Y(t^-)=q\}}$, $q \neq r \in \mathcal{S}$, and $\Lambda_{qr}(t) = \int_0^t \lambda_{qr}(s) ds$.

Under mild regularity conditions the assumptions allowing (1) are also satisfied by multivariate counting processes and hence by (not necessarily Markovian) multi-state processes, where each change of state can be represented as a jump in a multivariate counting process (cf. Andersen et al., 1993). This leads to the following more general definition of local independence.

**Definition 2.4** *Local independence*
Let $\mathbf{Y} \sim (Y_1, \ldots, Y_K)$ satisfy the above assumptions. Then, $Y_j$ is locally independent of $Y_k$, $k \neq j$, if and only if $\Lambda_j(t)$ is $\mathcal{F}_t^{-k}$–measurable for all $t \in \mathcal{T}$.

In words this means that the predictor $\Lambda_j$ does not depend on the past history of the component $Y_k$ given the whole past of all other components.

## 3 GRAPHICAL REPRESENTATION

We now formally define the graphs representing local independence structures.

**Definition 3.1** *Local independence graph*
Let $\mathbf{Y}_V \sim (Y_1, \ldots, Y_K)$ be a composable process satisfying the assumptions of Section 2.2 and let $G = (V, E)$ be a directed graph, where $V = \{1, \ldots, K\}$ and $E \subset \{(j,k) | j, k \in V, j \neq k\}$. Then $G$ is the local independence graph of $\mathbf{Y}_V$ if for all $j, k \in V$

$$(j, k) \notin E \quad \Leftrightarrow \quad \{j\} \not\to \{k\} \mid V \setminus \{j, k\}. \tag{2}$$

We can regard (2) as a dynamic version of the pairwise Markov property (Lauritzen, 1996) and hence call it the *pairwise dynamic Markov property*.

In the special case of $\mathbf{Y}_V$ being a homogenous Markov process its local independence graph is the same as a continuous time Bayesian network (Nodelman et al., 2002).

A local independence graph is directed but not necessarily acyclic. Furthermore, it allows two edges between two vertices, one in each direction. The notation familiar for DAGs can still be applied, such as parents, ancestors, non–descendants. In particular we will need the notion of an ancestral set $\mathrm{An}(A) = \mathrm{an}(A) \cup A$. Also, the operation of moralizing a graph can be carried out as usual, by marrying parents of a common child and then making the whole graph undirected, i.e. $G^m = (V, E^m)$ with $E^m = \{\{j, k\} | (j, k) \in E$ or $(k, j) \in E\} \cup \{\{j, k\} | \exists v \in V : j, k \in \mathrm{pa}(v)\}$. In $G^m$ we can check for usual separation in undirected graphs and will denote this by the symbol $\perp\!\!\!\perp_u$.

In Section 2.1 we said that the transition intensities can be summarized as $\alpha_k(t; (\mathbf{y}_{\mathrm{cl}(k)}, y'_k))$. In the general case this corresponds to the following property.

**Definition 3.2** *Local dynamic Markov property*
Let $G = (V, E)$ be a directed graph. For a multivariate process $\mathbf{Y}_V$ the property

$$\forall k \in V: \quad V \setminus \mathrm{cl}(k) \not\to \{k\} \mid \mathrm{pa}(k), \tag{3}$$

is called the *local dynamic Markov property w.r.t. G*.

In general (2) does not imply (3), e.g. if two components, $k, l$ say, contain the same information we might have that $\{k\} \not\to \{j\} | \{l\}$ as well as $\{l\} \not\to \{j\} | \{k\}$ without $\{k, l\} \not\to \{j\}$. However under additional conditions, such as the one of orthogonal martingales, we can show that in our setting the implication holds, because counting processes that are almost sure identical do not have orthogonal martingales (Andersen et al., 1993, p.73). The precise condition for the implication $(2) \Rightarrow (3)$ is given by (9) below.

In order to formulate a *global* dynamic Markov property we need a suitable notion of separation.

**Definition 3.3** *δ–separation*
Let $G = (V, E)$ be a directed graph. For $B \subset V$, let $G^B$ denote the graph given by deleting all directed edges of $G$ starting in $B$, i.e. $G^B = (V, E^B)$ with $E^B = E \setminus \{(j,k) | j \in B, k \in V\}$.
Then, we say for pairwise disjoint subsets $A, B, C \subset V$ that $C$ *δ–separates $A$ from $B$ in $G$* if $A \perp\!\!\!\perp_u B | C$ in the undirected graph $(G^B_{\text{An}(A \cup B \cup C)})^m$.
In case that $A, B$ and $C$ are not disjoint we define that $C$ δ–separates $A$ from $B$ if $C \setminus B$ δ–separates $A \setminus (B \cup C)$ from $B$. The empty set is always separated from $B$. Additionally, we define that the empty set δ–separates $A$ from $B$ if $A$ and $B$ are unconnected in $(G^B_{\text{An}(A \cup B)})^m$.

Note that deleting edges out of $B$ does not change $\text{An}(A \cup B \cup C)$ so that $G^B_{\text{An}(A \cup B \cup C)}$ is not ambiguous. Equivalently to Definition 3.3 we can use a trail condition, where a trail is a sequence of not necessarily direction preserving edges connecting two nodes. Define that any *allowed trail from $A$ to $B$* contains no edge of the form $(b, k), b \in B, k \in V \setminus B$. Then, for disjoint subsets $A, B, C$ of $V$, we have that $C$ δ–separates $A$ from $B$ if and only if all allowed trails from $A$ to $B$ are d–separated by $C$ (Didelez, 2000, pp.22).

As mentioned above, δ–separation is not symmetric in $A$ and $B$. This can be seen in the simple graph given in Figure 2(a). From the moral graphs in Figures 2(b) and (c) we see that $a \perp\!\!\!\perp_u b | c$ in $(G^a)^m$ but not in $(G^b)^m$. Alternatively, there are two trails between $a$ and $b$: $\{(a, b)\}$ and $\{(c, a), (b, c)\}$. Consider separating $b$ from $a$, then the first trail is not allowed and the second is d–separated by $c$ since the directed edges do not meet head–to–head in $c$. In contrast, if we want to separate $a$ from $b$, the second path is not allowed and the first is not d–separated by $c$. Hence $c$ δ–separates $b$ from $a$ but not $a$ from $b$.

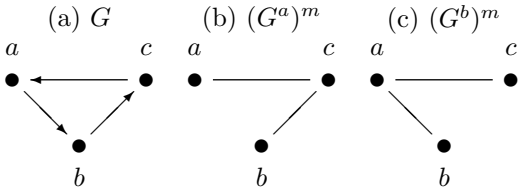

Figure 2: Example for the asymmetry of δ–separation.

The significance of δ–separation is due to the following global dynamic Markov property.

**Definition 3.4** *Global dynamic Markov property*
Let $\mathbf{Y}_V \sim (Y_1, \ldots, Y_K)$ be a composable process and $G = (V, E)$ a directed graph. For disjoint subsets $A, B, C \subset V$ the property

$$C \text{ δ–separates } A \text{ from } B \text{ in } G \Rightarrow A \not\to B \mid C. \quad (4)$$

is called the *global dynamic Markov property w.r.t. $G$*.

To understand why δ–separation works, consider again the above example, Figure 2(a). For a Markov process and small $h > 0$ this corresponds to the infinitesimal conditional independencies in Figure 3. Checking, e.g. whether $c$ δ–separates $b$ from $a$ corresponds to $Y_a(t + h) \perp\!\!\!\perp Y_b(t) | (Y_a(t), Y_c(t))$, which can be verified in Figure 3. Whereas $c$ δ–separating $a$ from $b$ would mean to $Y_a(t) \perp\!\!\!\perp Y_b(t + h) | (Y_b(t), Y_c(t))$, which is clearly not the case.

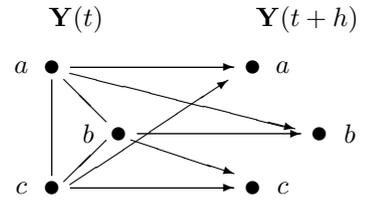

Figure 3: Infinitesimal dependence structure.

## 4 ASYMMETRIC GRAPHOIDS

The (semi)graphoid axioms first introduced by Dawid (1979) (cf. also Pearl and Paz, 1987; Pearl, 1988, p. 84; Dawid, 2001) are properties that we would like to be satisfied by an irrelevance relation. They hold for instance for undirected graph separation $\perp\!\!\!\perp_u$ and for d–separation (Verma and Pearl, 1988) as well as for probabilistic conditional independence. Symmetry is one of the basic (semi)graphoid properties. As yet, there is no general framework for asymmetric irrelevance relations available although other examples apart from local independence exist (e.g. Dawid, 1979, 1980; Galles and Pearl, 1996; Cozman and Walley, 1999). In the following we re–define the (semi)graphoid axioms for the asymmetric case and check which of them are satisfied by δ–separation and local independence.

**Definition 4.1** *Asymmetric (semi)graphoid*
Consider a lattice $(\mathcal{V}, \leq)$, where $A \vee B$ denotes the least upper bound and $A \wedge B$ the largest lower bound. Further, let $(A \text{ IR } B | C)$ be a ternary relation on this lattice. The following properties are called the *asymmetric semi–graphoid properties*:

*Left redundancy:* $A \text{ IR } B \mid A$

*Right redundancy:* $A \text{ IR } B \mid B$

*Left decomposition:* $A \text{ IR } B \mid C, D \leq A \Rightarrow D \text{ IR } B \mid C$

*Right decomposition:*
   $A \text{ IR } B \mid C,\ D \leq B \Rightarrow A \text{ IR } D \mid C$

*Left weak union:*
   $A \text{ IR } B \mid C,\ D \leq A \Rightarrow A \text{ IR } B \mid (C \vee D)$

*Right weak union:*
   $A \text{ IR } B \mid C,\ D \leq B \Rightarrow A \text{ IR } B \mid (C \vee D)$

*Left contraction:*
   $A \text{ IR } B \mid C \text{ and } D \text{ IR } B \mid (A \vee C) \Rightarrow (A \vee D) \text{ IR } B \mid C$

*Right contraction:*
   $A \text{ IR } B \mid C \text{ and } A \text{ IR } D \mid (B \vee C) \Rightarrow A \text{ IR } (B \vee D) \mid C$

If, in addition, the following properties hold we have an *asymmetric graphoid*:

*Left intersection:*
   $A \text{ IR } B \mid C \text{ and } C \text{ IR } B \mid A \Rightarrow (A \vee C) \text{ IR } B \mid (A \wedge C)$

*Right intersection:*
   $A \text{ IR } B \mid C \text{ and } A \text{ IR } C \mid B \Rightarrow A \text{ IR } (B \vee C) \mid (B \wedge C)$.

While Definition 4.1 applies for possibly overlapping sets, the following lemma clarifies the conditions under which it suffices to consider non–overlapping sets.

**Lemma 4.2** *Irrelevance for disjoint sets*
Let $\mathcal{V}$ be the power set of $V$ and $A, B, C \in \mathcal{V}$. For a ternary relation $A \text{ IR } B \mid C$ that satisfies left redundancy, decomposition, and contraction we have that

$$A \text{ IR } B \mid C \quad \Leftrightarrow \quad A \backslash C \text{ IR } B \mid C. \quad (5)$$

For a ternary relation that satisfies right redundancy, decomposition, and contraction we have that

$$A \text{ IR } B \mid C \quad \Leftrightarrow \quad A \text{ IR } B \backslash C \mid C. \quad (6)$$

**Proof:** To see (5) note that it follows directly from left decomposition that $A \text{ IR } B \mid C \Rightarrow A \backslash C \text{ IR } B \mid C$. To show $A \backslash C \text{ IR } B \mid C \Rightarrow A \text{ IR } B \mid C$, note that trivially $A \backslash C \text{ IR } B \mid (C \cup (C \cap A))$. Additionally, it follows from left redundancy (i.e. $C \text{ IR } B \mid C$) and left decomposition that $(C \cap A) \text{ IR } B \mid C$. Left contraction now yields the desired result. Property (6) is shown similarly.

The following corollary exploits (5) and (6) to reformulate the intersection property in a more familiar way.

**Corollary 4.3** *Alternative intersection property*
Let $\mathcal{V}$ be the power set of $V$ and $A, B, C \in \mathcal{V}$. Given an ternary relation with property (5), left decomposition, and left intersection. For pairwise disjoint sets $A, B, C, D \in \mathcal{V}$ it holds that

$$A \text{ IR } B \mid (C \cup D) \quad \text{and} \quad C \text{ IR } B \mid (A \cup D)$$
$$\Rightarrow \quad (A \cup C) \text{ IR } B \mid D. \quad (7)$$

With property (6), right decomposition, and right intersection it holds that

$$A \text{ IR } B \mid (C \cup D) \quad \text{and} \quad A \text{ IR } C \mid (B \cup D)$$
$$\Rightarrow \quad A \text{ IR } (B \cup C) \mid D. \quad (8)$$

**Proof:** With property (5) we have that $A \text{ IR } B \mid (C \cup D) \Leftrightarrow (A \cup C \cup D) \text{ IR } B \mid (C \cup D)$ from where it follows with left decomposition that $(A \cup D) \text{ IR } B \mid (C \cup D)$. With the same argument we get $C \text{ IR } B \mid (A \cup D) \Rightarrow (C \cup D) \text{ IR } B \mid (A \cup D)$. Left intersection yields $(A \cup C \cup D) \text{ IR } B \mid D$ which is again equivalent to $(A \cup C) \text{ IR } B \mid D$ because of (5). Implication (8) can be shown similarly.

## 4.1 PROPERTIES OF LOCAL INDEPENDENCE

An obvious way to translate local independence into an irrelevance relation is by letting $A \text{ IR } B \mid C$ stand for $A \not\rightarrow B \mid C$, where $A, B, C \subset V$. The semi–order is given by the set inclusion '$\subset$', the join and meet operations by union and intersection, respectively.

**Proposition 4.4** *Local independence as graphoid*
The following properties hold for local independence:

(i) left redundancy,

(ii) left decomposition,

(iii) left and right weak union,

(iv) left contraction,

(v) right intersection.

**Proof:** Didelez (2000, pp.70)

With the above proposition we have that (5) holds, i.e. $A \not\rightarrow B \mid C \Leftrightarrow A \backslash C \not\rightarrow B \mid C$. In contrast, it is clear by the definition of local independence that right redundancy does not hold because otherwise any process would always be locally independent of any other process given its own past. It follows that property (6) does not hold. Instead we have:

**Lemma 4.5** *Special version of right decomposition*
The following implication holds for local independence:

$$A \not\rightarrow B \mid C, \text{ and } D \subset B \quad \Rightarrow \quad A \not\rightarrow D \mid (C \cup B) \backslash D$$

The importance of intersection for the equivalence of pairwise, local and global Markov properties in undirected conditional independence graphs is well–known (cf. Lauritzen, 1996). It is of similar importance for local independence graphs.

**Proposition 4.6** *Left intersection*
Under the assumption that

$$\mathcal{F}_t^A \cap \mathcal{F}_t^B = \mathcal{F}_t^{A \cap B} \quad \forall\, A, B \subset V, \quad \forall\, t \in \mathcal{T} \qquad (9)$$

the property of left intersection holds for local independence.

**Proof:** Left intersection assumes that the $\mathcal{F}_t^{A \cup B \cup C}$–compensators $\Lambda_k(t)$, $k \in B$, are $\mathcal{F}_t^{B \cup C}$– as well as $\mathcal{F}_t^{A \cup B}$–measurable. With (9) we get that they are $\mathcal{F}_t^{B \cup (A \cap C)}$–measurable which yields the desired result.

Property (9) formalizes the intuitive notion that the considered components of the process are 'different enough' to ensure that common events in the past are necessarily due to common components. Its main consequence is the following.

**Corollary 4.7** *Pairwise implies local dynamic MP*
Under (9) the pairwise dynamic Markov property (2) and the local one (3) are equivalent.

**Proof:** (3) $\Rightarrow$ (2) follows from left weak union and left decomposition. (2) $\Rightarrow$ (3) follows immediately from (7) which holds if (9) holds.

Right decomposition requires special consideration because it makes a statement about the irrelevance of a subprocess $\mathbf{Y}_A$ after discarding part of the possibly *relevant* information $\mathbf{Y}_{B \setminus D}$. If the irrelevance of $\mathbf{Y}_A$ is due to knowing the past of $\mathbf{Y}_{B \setminus D}$ then it will not necessarily be irrelevant anymore if the latter is discarded. It therefore only holds under specific restrictions on the relation among the processes. The first restriction exploits property (9) to show that under an additional condition right decomposition also holds for local independence.

**Proposition 4.8** *Right decomposition (I)*
Given (9) right decomposition holds for local independence under the following additional conditions:

$$(B \cap A) \setminus (C \cup D) = \emptyset \quad \text{and} \quad B \not\to D \mid A \cup C. \qquad (10)$$

**Proof:** Didelez (2000, p.72)

Another situation where right decomposition holds is given as follows.

**Proposition 4.9** *Right decomposition (II)*
Let $A, B, C, D \subset V$ with $(B \cap A) \setminus (C \cup D) = \emptyset$. Right decomposition holds under the conditions that

$$B \not\to A \setminus (C \cup D) \mid (C \cup D) \qquad (11)$$

and

$$A \not\to \{k\} \mid C \cup B \quad \text{or} \quad B \not\to \{k\} \mid (C \cup D \cup A) \qquad (12)$$

for all $k \in C \setminus D$.

**Proof:** Didelez (2000, pp.72)

## 4.2 PROPERTIES OF $\delta$–SEPARATION

The interpretation of $\delta$–separation as irrelevance relation should be that if $C$ $\delta$–separates $A$ from $B$ in $G$ then $A$ is irrelevant for $B$ given $C$. This is denoted by $A \operatorname{IR}_\delta B | C$.

**Proposition 4.10** *$\delta$–separation as graphoid*
$\delta$–separation satisfies the following properties:

(i) left redundancy,

(ii) left decomposition,

(iii) left and right weak union,

(iv) left and right contraction,

(v) left and right intersection.

**Proof:** Didelez (2000, pp.27).

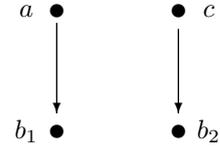

Figure 4: Counterexample for property (6).

With these results not only property (5) but also (7) hold for $\delta$–separation. So far, however, the conditions for (6) and (8) are not satisfied because right redundancy does not hold. By definition right redundancy would imply that $A \operatorname{IR}_\delta B | B \Leftrightarrow A \setminus B \operatorname{IR}_\delta B | \emptyset$ which is only true if $A \setminus B$ and $B$ are unconnected in $(G^B_{\operatorname{An}(A \cup B)})^m$. A simple counterexample is given by the graph with $V = \{a, b\}$ and $E = \{(a, b)\}$. Additionally, property (6) does not hold, as can be seen by another example shown in Figure 4. Let $A = \{a\}$, $B = \{b_1, b_2\}$, and $C = \{b_1, c\}$. Then, $A \operatorname{IR}_\delta B \setminus C | C$ but not $A \operatorname{IR}_\delta B | C$ since the latter only holds if $A \operatorname{IR}_\delta B | C \setminus B$. In contrast, the converse does hold because it is a special case of the subsequent result paralleling Lemma 4.5.

**Lemma 4.11** *Special version of right decomposition*
Given a directed graph $G$, it holds that:

$$A \operatorname{IR}_\delta B \mid C,\ D \subset B \Rightarrow A \operatorname{IR}_\delta D \mid (C \cup B) \setminus D$$

**Proof:** Let $A^* = A \setminus (B \cup C)$ and $C^* = C \setminus B$. Then, we have to show that $A^* \perp\!\!\!\perp_u B | C^*$ in $(G^B_{\operatorname{An}(A \cup B \cup C)})^m$ implies $A^* \perp\!\!\!\perp_u D | C^* \cup (B \setminus D)$ in $(G^D_{\operatorname{An}(A \cup B \cup C)})^m$. Note that $A^* \perp\!\!\!\perp_u D | C^* \cup (B \setminus D)$ in $(G^B_{\operatorname{An}(A \cup B \cup C)})^m$ holds due to weak union and decomposition of $\perp\!\!\!\perp_u$. Changing the graph to $(G^D_{\operatorname{An}(A \cup B \cup C)})^m$ means that all edges that are present in $G$ as directed edges starting in $B\backslash D$ are added. Additionally, those edges have to be added which result from vertices in $B\backslash D$ having common children with other vertices. Since all these new edges involve vertices in $B\backslash D$ there can be no additional path between $A^*$ and $D$ in $(G^D_{\text{An}(A\cup B\cup C)})^m$ not intersected by $C^*\cup(B\backslash D)$.

Although (6) does not hold in full generality it is easily checked that (8) does.

**Proposition 4.12** *Alternative intersection property*
Property (8) holds for $\delta$–separation, i.e.

$$A\,\text{IR}_\delta\,B\mid(C\cup D) \quad \text{and} \quad A\,\text{IR}_\delta\,C\mid(B\cup D)$$
$$\Rightarrow \quad A\,\text{IR}_\delta\,(B\cup C)\mid D$$

for pairwise disjoint sets $A,B,C,D$.

**Proof:** Given that $A\perp\!\!\!\perp_u B|(C\cup D)$ in $(G^B_{A\cup B\cup C\cup D})^m$ and $A\perp\!\!\!\perp_u C|(B\cup D)$ in $(G^C_{A\cup B\cup C\cup D})^m$, both separations also hold in $(G^{B\cup C}_{A\cup B\cup C\cup D})^m$. With the properties of $\perp\!\!\!\perp_u$ it follows that $A\perp\!\!\!\perp_u (B\cup C)|D$ in $(G^{B\cup C}_{A\cup B\cup C\cup D})^m$.

The above proposition does not necessarily hold if $B,C,D$ are not disjoint. But it can be shown by a very similar proof that it remains valid if $A\cap B\neq\emptyset$ or $A\cap C\neq\emptyset$.

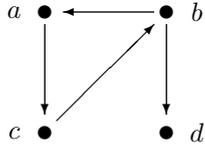

Figure 5: Counterexample for right decomposition.

A simple counterexample for right decomposition of $\delta$–separation is given in Figure 5, where $\{c\}$ $\delta$–separates $\{a\}$ from $B=\{b,d\}$ but it is not true that $\{c\}$ $\delta$–separates $\{a\}$ from $\{d\}$. In the first case we delete the arrow from $b$ to $a$ whereas in the second case it is kept. However, we can obtain the following result paralleling Propositions 4.8 and 4.9.

**Proposition 4.13** *Right decomposition*
Right decomposition holds for $\delta$–separation in the special case that $(A\cap B)\backslash(C\cup D)=\emptyset$ and

(i) either $B\,\text{IR}_\delta\,D|(A\cup C)$

(ii) or $B\,\text{IR}_\delta\,A\backslash(C\cup D)|(C\cup D)$ and for all $k\in C\backslash D$ either $A\,\text{IR}_\delta\,\{k\}|((C\backslash\{k\})\cup B)$ or $B\,\text{IR}_\delta\,\{k\}|((C\backslash\{k\})\cup D\cup A)$.

**Proof:** Didelez (2000, pp.31).

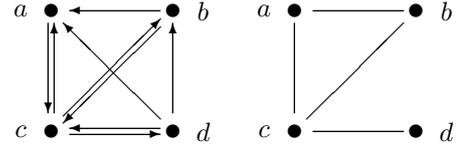

Figure 6: Illustration of condition (i).

Property (i) of Proposition 4.13 is illustrated in Figure 6 which includes the moral graph $(G^d)^m$. Choose $A=\{a\}$, $B=\{b,d\}$ and $C=\{c\}$. Property (ii) of Proposition 4.13 is illustrated in Figure 7 which includes the moral graph $(G^d)^m$. Choose $A=\{a\}$, $B=\{b,d\}$ and $C=\{c\}$. Roughly one can summarize the assumptions of Proposition 4.13 as that either $B\backslash D$ does not affect $D$ or $A$ and $B\backslash D$ do not affect common nodes in $C$ as this would open a path between them.

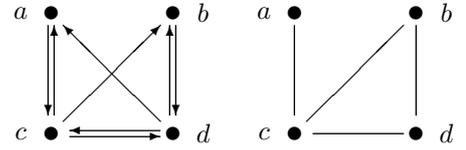

Figure 7: Illustration of condition (ii).

The main consequence of the properties of local independence and $\delta$–separation is that we can indeed use $\delta$–separation to read off local independencies among sub–processes from a local independence graph.

**Theorem 4.14** *Pairwise implies global MP*
With the above properties, using right decomposition under the conditions specified by Propositions 4.8, 4.9 and 4.13, we have that the pairwise dynamic Markov property (2) and the global dynamic Markov property (4) are equivalent.

**Proof:** That (4) implies (2) is easily seen because $V\backslash\{j,k\}$ $\delta$–separates $j$ from $k$ if $(j,k)\notin E$. The structure of the proof of $(2)\Rightarrow(4)$ corresponds to the one given by Lauritzen (1996, p. 34) for the equivalence of the Markov properties in undirected conditional independence graphs. Due to the asymmetry of local independence the present proof is more complicated and is given in Didelez (2000, pp.90).

In order to illustrate the significance of Theorem 4.14 we return to the example given in the Introduction and depicted in Figure 1. We can see from the graph that in an analysis that only uses data on the visits, hospitalization and survival, i.e. where health status is ignored, we might find that survival depends on the visits. This is somewhat surprising as the visit process is locally independent of the health status process given the other components. The reason is that the node 'visits' is not separated from 'survival' by 'hospitalization' alone. In other words, a history of e.g. hospitalization without prior visit by a nurse carries a different information than a history of hospitalization with prior visit — it is informative for the unobserved health process. The health status could be regarded as a particular type of time dependent confounder.

# 5 DISCUSSION

Local independence graphs and $\delta$–separation represent dynamic dependencies among continuous time multi–state processes. They are a valuable tool for reasoning in dynamic settings in the presence of time–dependent confounding and censoring (Didelez, 2005a). In addition, the framework of asymmetric (semi)graphoids developed here can be applied more generally to asymmetric irrelevance relations.

Similar to the classical Bayesian networks, the graph structure can be exploited to simplify and accelerate computations. In this context some further results on collapsibility and likelihood factorization are derived in Didelez (2005a, 2005b). Such a likelihood factorization is also exploited by Nodelman et al. (2003) for learning the graph. However, as can be seen from their work the computational side is more intricate than for standard networks and deserves further investigation.

**Acknowledgements**

This work has been supported by the German Research Foundation (SFB 386) and the Centre for Advanced Study at the Norwegian Academy of Science and Letters.

**References**

Aalen, O.O. (1987). Dynamic modelling and causality. *Scand. Actuar. J.,* 177-90.

Andersen, P.K., Borgan,Ø. Gill, R.D. & Keiding, N. (1993). *Statistical models based on counting processes.* Springer, New York.

Cozman F.G. and Walley, P. (1999). Graphoid properties of epistemic irrelevance and independence. Technical Report, University of Sao Paolo.

Dawid, A.P. (1979). Conditional independence in statistical theory. *J. Roy. Statist. Soc. Ser. B* **41**, 1-31.

Dawid, A.P. (1980). Conditional independence for statistical operations. *The Annals of Statistics*, **8**, 598-617.

Dawid, A.P. (2001). Separoids: A mathematical framework for conditional independence and irrelevance. *Ann. Math. Artificial Intelligence*, **32**, 335–72.

Didelez, V. (2000). *Graphical models for event history data based on local independence.* Logos, Berlin.

Didelez, V. (2005a). Graphical models for marked point processes based on local independence. *Tech. Rep. No. 258,* University College London.

Didelez, V. (2005b). Graphical models for composable finite Markov processes. *Scandinavian Journal of Statistics*, to appear.

Eichler, M. (2002). Granger-causality and path diagrams for multivariate time series. *Journal of Econometrics*, to appear.

Galles, D. and Pearl, J. (1996). Axioms of causal relevance. Technical Report 240, University of California, Los Angeles.

Granger, C.W.J. (1969). Investigating causal relations by econometric models and cross–spectral methods. *Econometrica* **37**, 424-38.

Lauritzen, S.L. (1996). *Graphical models.* Clarendon Press, Oxford.

Nodelman, U., Shelton, C. R. & Koller, D. (2002). Continuous time Bayesian networks. *Proceedings of the 18th Conference on Uncertainty in Artificial Intelligence*, 378-87.

Nodelman, U., Shelton, C. R. & Koller, D. (2003). Learning continuous time Bayesian networks. *Proceedings of the 19th Conference on Uncertainty in Artificial Intelligence*, 451-8.

Pearl, J. (1988). *Probabilistic reasoning in intelligent systems.* Morgan Kaufmann, San Mateo.

Pearl, J., Dechter, R. (1996). Identifying independencies in causal graphs with feedback. *Proceedings of the 12th Conference on Uncertainty in Artificial Intelligence*, 240-6.

Pearl, J. & Paz, A. (1987). Graphoids: A graph–based logic for reasoning about relevancy relations. In B. Du Boulay, D. Hogg, & L. Steel (Eds.), *Advances in Artificial Intelligence – II*, pp. 357-63.

Schweder, T. (1970). Composable Markov processes. *J. Appl. Prob.* **7**, 400-10.

Spirtes, P. (1995). Directed cyclic graphical representations of feedback models. *Proceedings of the 11th Conference on Uncertainty in Artificial Intelligence*, 491-8.

Verma, T. & Pearl, J. (1988). Causal networks: Semantics and expressiveness. *Proceedings of the 4th Conference on Uncertainty in Artificial Intelligence*, 69-76.